\documentclass[]{interact}

\usepackage{epstopdf}
\usepackage{subfigure}

\usepackage{natbib}
\bibpunct[, ]{(}{)}{;}{a}{}{,}
\renewcommand\bibfont{\fontsize{10}{12}\selectfont}

\theoremstyle{plain}

\theoremstyle{definition}

\theoremstyle{remark}

\usepackage{algorithm}
\usepackage{algorithmic}
\usepackage{multirow}
\usepackage{booktabs}
\usepackage{amsmath}
\usepackage[dvipsnames]{xcolor}
\usepackage{lineno}

\begin{document}


\title{Semi-Supervised Contrastive Learning for Remote Sensing: Identifying Ancient Urbanization in the South-Central Andes}
\author{
\name{Jiachen Xu\textsuperscript{a$\ast$}, Junlin Guo\textsuperscript{b$\ast$}, James Zimmer-Dauphinee\textsuperscript{c}, Quan Liu\textsuperscript{b}, Yuxuan Shi\textsuperscript{a}, Zuhayr Asad\textsuperscript{b}, D. Mitchell. Wilkes\textsuperscript{b}, Parker VanValkenburgh\textsuperscript{d}, Steven A. Wernke\textsuperscript{c}, Yuankai Huo\textsuperscript{b}\thanks{\textsuperscript{$^\ast$}First Co-author. Contribute equally to this paper.}\thanks{CONTACT Yuankai Huo. Email: yuankai.huo@vanderbilt.edu}}
\affil{\textsuperscript{a}School of Engineering, Vanderbilt University, Nashville, USA; \textsuperscript{b}Department of Electrical Engineering and Computer Science, Vanderbilt University, Nashville, USA;\textsuperscript{c}Department of Anthropology, Vanderbilt University, Nashville, USA;\textsuperscript{d}Department of Anthropology, Brown University, Providence, USA}
}

\maketitle

\begin{abstract}
Archaeology has long faced fundamental issues of sampling and scalar representation. Traditionally, the local-to-regional-scale views of settlement patterns are produced through systematic pedestrian surveys. Recently, systematic manual survey of satellite and aerial imagery has enabled continuous distributional views of archaeological phenomena at interregional scales. However, such “brute force” manual imagery survey methods are both time- and labor-intensive, as well as prone to inter-observer differences in sensitivity and specificity. The development of self-supervised learning methods (e.g., contrastive learning) offers a scalable learning scheme for locating archaeological features using unlabeled satellite and historical aerial images. However, archaeological features are generally only visible in a very small proportion relative to the landscape, while the modern contrastive-supervised learning approach typically yields an inferior performance on highly imbalanced datasets. In this work, we propose a framework to address this long-tail problem. As opposed to the existing contrastive learning approaches that typically treat the labeled and unlabeled data separately, our proposed method reforms the learning paradigm under a semi-supervised setting in order to fully utilize the precious annotated data (\textless 7\% in our setting). Specifically, the highly unbalanced nature of the data is employed as the prior knowledge in order to form pseudo negative pairs by ranking the similarities between unannotated image patches and annotated anchor images. In this study, we used 95,358 unlabeled images and 5,830 labeled images in order to solve the issues associated with detecting ancient buildings from a long-tailed satellite image dataset. From the results, our semi-supervised contrastive learning model achieved a promising testing balanced accuracy of 79.0\%, which is a 3.8\% improvement as compared to other state-of-the-art approaches.
\end{abstract}

\begin{keywords}
Machine Learning; Satellite Imagery; Semi-Supervised Learning; Contrastive Learning
\end{keywords}

\section{Introduction}
Archaeological structures and settlements are essential sources of information that archaeologists use to study the economic, political, and social systems of ancient civilizations. Conventional approaches to mapping and recording settlement locations at local and regional scales have relied on field-based pedestrian survey methods, which require professionals to physically examine the landscape for evidence of ancient material culture~\citep{banning2002archaeological,phillips1953method,sanders1961developmental,balkansky2000archaeological,alcock2016side}. However, the scale of field-based surveys is ultimately limited by the physical impedances of fieldwork. Moreover, the distribution of both survey and excavation zones is often unsystematic, which further complicates efforts to synthesize findings across field projects. Since the early 2000s, archaeologists have made use of high-resolution satellite imagery to understand the spatial and structural patterns of the archaeological features at larger scales, including by step-wise visual identification of sites by trained specialists~\citep{hanson2012archaeology,fowler2002satellite,lasaponara2012satellite, bewley2016endangered, casana2014regional, lin2014crowdsourcing, parcak_archaeology_2019}. Such research has produced novel insights into macro- and inter-regional scale settlement patterns~\citep{casana2014regional,casana2013corona,wernke2020interregional}. However, such “brute force” manual imagery survey methods are very labor intensive, time-consuming, and prone to inter-observer differences in feature detection sensitivity and specificity~\citep{casana2014regional}. In part, these issues are inherent and can be attributed to the nature of the data, as archaeological features are generally very sparsely distributed across the landscape making manual identification and the labeling of archaeological features in satellite imagery a very low yield endeavor. Observational fatigue and inter-observer differences in detection rates additionally pose unavoidable risks. The resulting datasets are thus generally quite large in aerial extent but come with few labels~\citep{mnih2012learning,casana2013corona}. Developing an effective machine learning algorithm for automating information extraction procedures on such large-scale, sparsely annotated, and unbalanced data is a long-standing machine-learning challenge in remote sensing.

In recent years, the rapid development of self-supervised contrastive learning has shown promise towards the task of utilizing large-scale, sparsely annotated data. However, the proportion of images containing archaeological settlements is often relatively low (\textless 7\% in our setting). Such an unbalanced data distribution is problematic for modern contrastive learning algorithms~\citep{chen2021exploring,grill2020bootstrap}, consequently leading to excessive favoring representations of the majority classes. By reforming contrastive learning in a semi-supervised setting, we emphasize the critical role of the sparse but valuable annotated positive instances in both training and fine-tuning stages.

In this work, we propose a novel self-supervised contrastive learning framework to identify ancient settlements through relict architectural feature detection in the south-central Andes. As opposed to existing self-supervised learning approaches, which typically model labeled and unlabeled data separately, we introduce a holistic, end-to-end semi-supervised learning framework that utilizes the highly unbalanced nature of the data to form pseudo-negative pairs by ranking the similarities between unannotated image patches and annotated anchor images. Specifically, pseudo-negative images are employed to calculate a supervised contrastive (SupCon) loss~\citep{khosla2020supervised}, which is seamlessly integrated with the contrastive loss~\citep{chen2020simple}. 

\begin{figure}[h]
\begin{center}
\includegraphics[width=0.7\textwidth]{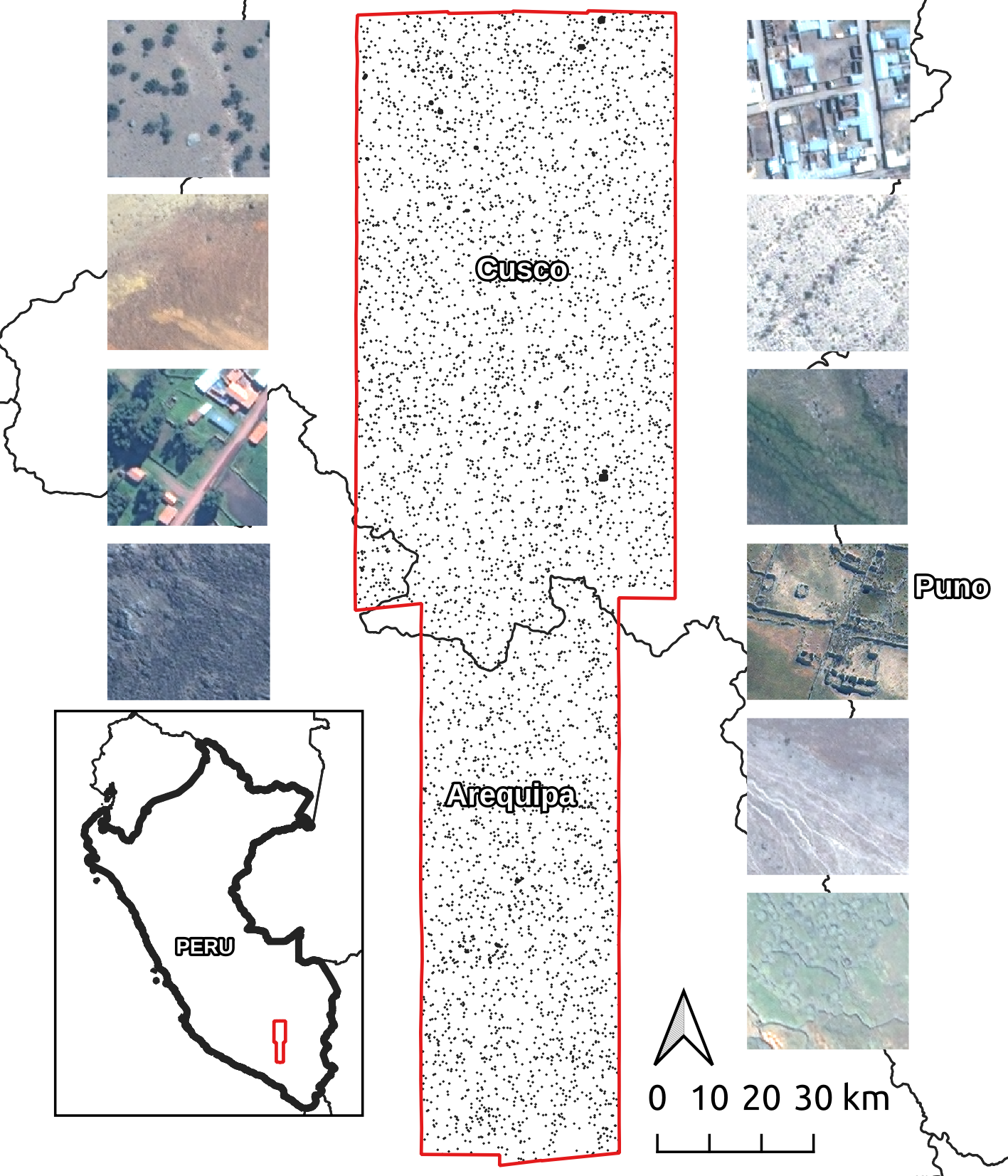}
\end{center}
\caption{\textbf{Survey Region.} The study region encompasses approximately 4,000 km$^2$ of the western cordillera of the southern Peruvian highlands, including portions of the modern Cusco and Arequipa districts. Sample tiles represent the diversity of land formation in the region and the sample locations are shown in black.}
\label{fig:overview}
\end{figure}

To test this approach, this project surveys an approximately 4,000 km$^2$ region of the western cordillera of the southern Peruvian Andes (Figure~\ref{fig:overview}). Utilizing images taken by Worldview 2 and Worldview 3 Satellite platforms, our dataset consists of 95,358 unlabeled images and 5,830 labeled images, where the ratio between positive and negative instances is roughly 1:100. We show that our semi-supervised contrastive learning model outperforms its self-supervised and fully-supervised counterparts, along with traditional supervised networks such as ResNet50.

There also have been recent general computer vision studies for semi-supervised contrastive learning methods~\citep{zhang2022semi, yang2022class} that allow for the optimal utilization of vast amounts of unlabeled data. These approaches tend to refine the quality of pseudo labels by continuously selecting positive samples. However, the primary challenge in archaeological field lies in the extremely imbalanced proportion of positive and negative samples in the archeological dataset (1:100). The highly imbalanced proportion of negative classes can lead to ineffective training in conventional self- and semi-supervised contrastive learning. In comparison, our proposed model offers several advantages:\textbf{(1)} Leveraging the foreground image, aligning with the class-specific few-shot learning design for the self-supervised contrastive task. \textbf{(2)} The supervised contrastive task is balanced and further ensures the discriminative representation between the two classes in the latent space.
\\\\
\textbf{Innovation of the work} 
\\
\\The innovation of this study is four-fold:
\begin{itemize}
    \item This study investigates a large, new survey region utilizing a cutting-edge representative deep learning approach. The study region encompasses approximately 4,000 km$^2$ of the western cordillera of the southern Peruvian highlands, including portions of the modern Cusco and Arequipa districts.
    \item We propose a novel contrastive learning scheme which is optimized for the unique challenges faced in remote sensing image analyses, such as \textbf{(1)} effectively learning design from limited annotated data with large-scale unannotated data, and \textbf{(2)} the highly imbalanced data distribution (e.g., the foreground objects of interests are much less than the background).
    \item A new semi-supervised contrastive learning method was introduced by aggregating the advantages from previous \textbf{(1)} self-supervised and \textbf{(2)} supervised contrastive learning strategies. Compared with traditional approaches, the proposed method maximizes utilization of large-scale unlabeled image data and small-scale labeled image data under a probabilistic learning model.
    \item A similarity-based down-sampling approach is proposed for pseudo-label synthesis in both the latent space learning section, as well as the supervised learning section, of the semi-supervised model.
\end{itemize}

\section{Background and Related Research}
This section provides an overview of the background and related research for contrastive representation learning and satellite remote sensing. The following brief literature survey includes a summary of recent contrastive learning methods and a discussion of the applications of machine learning in remote sensing research. 

\subsection{Contrastive Representation Learning}
In contrast to supervised learning~\citep{cunningham2008supervised}, which requires the presence of labeled inputs to predict outputs, self-supervised learning~\citep{le2013building} refers to the identification of the hidden patterns of a dataset without the usage of any labels. Comprising what is a relatively new family of self-supervised learning methods, contrastive representation learning has recently become a key approach in solving various computer vision tasks with state-of-the-art performance~\citep{wu2018unsupervised,noroozi2016unsupervised,zhuang2019local,hjelm2018learning,chuang2020debiased,tian2020makes,khosla2020supervised,cui2021parametric}. Designed to learn the general features of large datasets without labels, contrastive learning aims to pull similar sample pairs together while pushing dissimilar pairs apart. As a result, the model is capable of learning the high-level features of a dataset even with few or no labels available.

In recent years, various contrastive representation learning methods have been proposed with different implementations. SimCLR~\citep{chen2020simple} aims to pull the representations of different views of the same image closer while repulsing the views of different images in the latent space.SwAV~\citep{caron2020unsupervised} applies online clustering on different augmentations of the same image instead of performing explicit pairwise feature comparisons. Wu et al.~\citep{wu2018unsupervised} proposes the use of an offline memory bank to store all data representations, with training data randomly selected for negative-pair minimization. Instead of utilizing an offline dictionary, MoCo~\citep{he2020momentum} utilizes a momentum design to build a dynamic dictionary that stores a negative sample pool, which demands a large batch size. To further alleviate the cost of storing negative pairs, BYOL~\citep{grill2020bootstrap} is proposed to incorporate an asynchronous momentum encoder into the model so that it can use only the positive pairs for training. Recently, SimSiam~\citep{chen2021exploring} has been proposed to save GPU memory consumption by fully eliminating the momentum encoder. In addition, various efforts have been made to modify the contrastive learning approach within a fully-supervised setting; an example would be the SupCon loss proposed by Khosla~\citep{khosla2020supervised}.

\subsection{Remote Sensing with Machine Learning}
Satellite remote sensing has contributed to the execution of a variety of tasks, including climate change measurement, crop condition monitoring, natural disaster alerts, and archaeological site detection~\citep{harris1987satellite}. Satellites were first introduced to the field of archaeology in the late 1900s, with Landsat and SPOT imagery being used for archaeological predictive modeling and archaeological feature detection~\citep{leisz2013}. Since then, the usage of satellite remote sensing in the detection of archaeological sites has picked up rapidly, leveraging all available technologies, from decommissioned CORONA imagery~\citep{urCORONA2013} to the latest in multi-spectral imagery~\citep{abrams2013}.

\textcolor{black}{Starting in the 2000s, the development of machine learning (and more particularly, representation learning) offered major breakthroughs towards the analytical approaches employed on satellite images; this in turn lead to seminal insights and discoveries in the field of archaeology~\citep{lary2016machine,camps2009machine,ali2015review,cooner2016detection,comer_mapping_2013,parcak_archaeology_2019}. Depending on the specific problem, various types of machine learning algorithms have been employed, such as support vector machines (SVM), decision trees, random forests, etc.~\citep{samui2008support,azamathulla2011support,friedl1997decision,pal2005random}. Recently, many deep learning (and more specifically, contrastive learning-based) methods have been utilized towards remote sensing applications~\citep{hou2021hyperspectral, yue2021self, wang2022self, liu2020deep, hu2021contrastive}.} 

\section{Methods}
The overall design of our framework is shown in Figure~\ref{fig:framework}. An analysis of the backbone network and pseudo-label synthesis is presented below.

\begin{figure*}[ht]
\begin{center}
\includegraphics[width=1\textwidth]{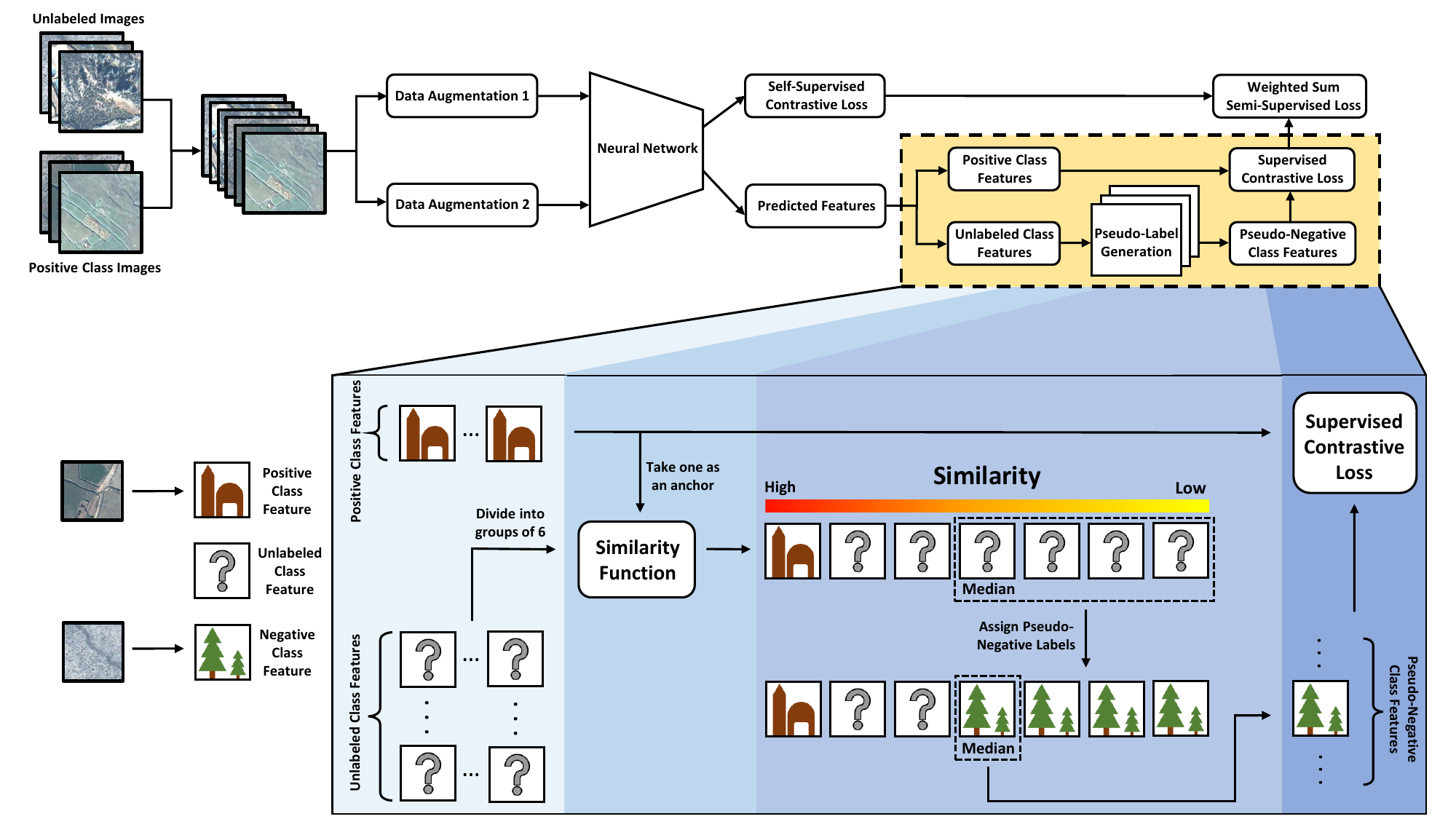}
\end{center}
\caption{\textbf{Overall framework.} This figure demonstrates the general structure of our semi-supervised contrastive learning framework. The upper panel shows the general flow of our framework, which was adopted from the SimSiam network. The lower panel describes the process of obtaining the supervised contrastive loss from predicted features. The detailed discussions can be found in the Methods section.}
\label{fig:framework}
\end{figure*}
\textcolor{black}{\subsection{Overview of the SimSiam Framework}} 
In this work, SimSiam is chosen as our backbone network due to its simplicity and effectiveness. Compared to other widely-used self-supervised representation learning networks, SimSiam removes all additional structures such as negative samples (SimCLR), momentum encoder (BYOL), or clustering (SwAV) and still obtains great performance in its learning representations of unlabeled datasets. \textcolor{black}{The overall loss function consists of two separate losses, namely (1) self-supervised contrastive loss (i.e., Cosine similarity) and (2) supervised contrastive loss. The rationale of employing both self-supervised and supervised loss is to form a new “semi-supervised” learning scheme for remote sensing image learning. Compared with traditional self-supervised contrastive learning~\citep{chen2021exploring} and supervised contrastive learning~\citep{khosla2020supervised} approaches, the proposed method maximizes utilization of large-scale unlabeled image data by incorporating the small-scale labeled image data.} In addition, the recently introduced mixed-precision training feature is utilized to accelerate the training process. 

\begin{algorithm}[ht]
\caption{Pseudo-Code for generating pseudo negative pairs}
\label{alg:algorithm}
\textbf{Input:} An array of unlabeled class features: $f\_un$ \\
\textbf{Input:} An array of positive class features: $f\_pos$ \\
\textbf{Output:} An array of pseudo-negative class features: $f\_neg$

\begin{algorithmic}[1] 
\STATE $f\_un = Normalize(f\_un)$
\STATE $f\_pos = Normalize(f\_pos)$
\STATE Divide $f\_un$ into groups of 16 $\rightarrow f\_un\_groups$
\STATE Randomly choose a feature from $f\_pos$ $\rightarrow f\_positive$
\STATE $f\_neg = [\ ]$
\FOR{$f\_un\_group$ in $f\_un\_groups$}
\STATE $sim\_array = Similarity\_Function(f\_un\_group + f\_positive)$
\STATE $f\_neg.append(Median(sim\_array))$
\ENDFOR
\STATE \textbf{return} $f\_neg$
\end{algorithmic}
\end{algorithm}
\subsection{Pseudo-Label Synthesis}
For the synthesis of pseudo-labels, a mixed array of unlabeled and positive class features is used as the starting point. The first step is to normalize this array. Next, the array is decoupled into $X$ features of unlabeled images (Group 1) and $Y$ features of positively-labeled images (Group 2). Then, we divide the $X$ unlabeled class features into subgroups of size $k$ (16 in our case), and we end up with $X/k$ subgroups. Meanwhile, a single feature is randomly selected from the $Y$ positive class features for future use.

For each subgroup, we apply the cosine similarity function in order to compute the similarity between the previously selected positive feature and the features in the subgroup. Following this, the unlabeled image with the median similarity score is assigned a pseudo label (i.e., negative class) based on the hypothesis that negative images dominate the distribution of the entire cohort. \textcolor{black}{The 1:100 ratio used in this paper follows the design from a previous publication~\citep{yang2020rethinking}. According to~\citep{yang2020rethinking}, a higher imbalance ratio can impose an additional challenge towards the classification tasks as compared to a scenario with moderately imbalanced data.} 

\textcolor{black}{The pseudo-code for generating such pseudo-labels is presented in Algorithm \ref{alg:algorithm}. As an example, it assumes that there are $N$ unlabeled images. The ratio of positive images ($N_{pos}$) to negative images ($N - N_{pos}$) is roughly 1:100 in this study. Thus, if the size of the batch is $B\ (B\ll N)$ and the images are randomly selected, the probability of having exactly $n$ positive image(s) in this batch is expressed as:}
\textcolor{black}{
\begin{equation}
p\left(n\right)=\frac{\binom{N_{pos}}{n} \cdot \binom{N - N_{pos}}{B-n}}{\binom{N}{B}}
\end{equation}}
\textcolor{black}{Following this, the probability of having one or less positive images (in other words,} \textcolor{black}{$B$ or $B-1$ negative images) among a randomly selected batch B, is:
\begin{equation}
p\left(n \leq 1\right)=\sum_{n=0}^{1} \frac{\binom{N_{pos}}{n} \cdot \binom{N - N_{pos}}{B-n}}{\binom{N}{B}} = \frac{\binom{N - N_{pos}}{B} + N_{pos}\cdot \binom{N - N_{pos}}{B-1}}{\binom{N}{B}} \approx 1 
\end{equation} 
when $N \approx N - N_{pos}$. Therefore, every batch contains almost all negative images, which ensures that the pseudo-labels are negative.}

\subsection{Semi-Supervised Contrastive Learning}
The key innovation of our method is our proposal of a new semi-supervised contrastive learning strategy. In short, we aggregate the standard cosine similarity loss with the supervised contrastive (SupCon) loss. 
\subsubsection{Self-supervised Contrastive Task}
In each iteration of the training, $X$ unlabeled images and $Y$ positively-labeled images are utilized as inputs for our SimSiam model, and the generated symmetric loss is named \textbf{loss\_cosine}. The formulas are defined as~\citep{chen2021exploring}: 
\begin{equation}
\mathcal{D}\left(p_{1}, z_{2}\right)=-\frac{p_{1}}{\left\|p_{1}\right\|_{2}} \cdot \frac{z_{2}}{\left\|z_{2}\right\|_{2}}
\end{equation}

\begin{equation}
\mathcal{L}=\frac{1}{2} \mathcal{D}\left(p_{1}, z_{2}\right)+\frac{1}{2} \mathcal{D}\left(p_{2}, z_{1}\right)
\end{equation}
\textcolor{black}{The formulas above show the SimSiam symmetrized loss for a single data point (image) $x$. $z_{1}$ and $z_{2}$ denote the encoding vectors of the two augmented views $x_{1}$ and $x_{2}$ (generated from $x$). Following, $p_{1}$ and $p_{2}$ denote the projection views of the encoding vectors $z_{1}$ and $z_{2}$ by adding an MLP head on top of the shared encoder. Equation (3) represents the negative cosine similarity. Equation (4) shows how a contrastive pair is generated through two augmented views and used for computing the cosine similarity loss.}

\subsubsection{Supervised Contrastive Task}
Following the self-supervised contrastive task, the encoding features, $z$ (that are mentioned in the self-supervised task), are employed to calculate the supervised loss, which is named \textbf{loss\_super}. The encoding features will go through the steps in the Pseudo-Label Synthesis subsection, and generate  $X/k$ pseudo-negative class features and $Y$ positive class features (mentioned earlier in the subsection). Finally, these features are combined together as inputs to the SupCon~\citep{khosla2020supervised} loss function in order to calculate loss\_super.

The formula for the SupCon loss is presented below~\citep{khosla2020supervised}:
\begin{equation}
\begin{split}
    \mathcal{L}_{\text {out }}^{s u p} & =\sum_{i \in I} \mathcal{L}_{\text {out }, i}^{s u p} \\
    & = \sum_{i \in I} \frac{-1}{|P(i)|} \sum_{p \in P(i)} \log \frac{\exp \left(\boldsymbol{z}_{i} \cdot \boldsymbol{z}_{p} / \tau\right)}{\sum_{a \in A(i)} \exp \left(\boldsymbol{z}_{i} \cdot \boldsymbol{z}_{a} / \tau\right)}
\end{split}
\end{equation}
\textcolor{black}{Equation (5) presents a generalized form of the supervised contrastive loss within a multi-view batch, where $i\in I \equiv \{1\ ... \ 2N \}$ is the index of an arbitrary augmented sample (view). The "$\cdot$" symbol denotes the dot product. The index $i$ represents the anchor image, while $P(i)$ represents all positive pairs of the anchor. $A(i) \equiv I\setminus \{i\}$, indicates the remaining $2N - 1$ views in the batch, excluding the anchor $i$. As shown in Equation (5), the contrastive nominator maximizes the similarity between the anchor and its positive pairs. Meanwhile, the contrastive denominator differentiates the anchor from negative samples.}

\subsubsection{Semi-supervised multi-task loss}
\textcolor{black}{In the final step, the total loss function is modeled as a multi-task loss design with different weighting parameters on the self-supervised (\textbf{loss\_cosine}) and supervised (\textbf{loss\_super}) losses. The optimization of the weighting parameters of the loss function is inspired by Kendall~\citep{kendall2018multi}, in which the weighting parameters ($v_1$ and $v_2$ in Equation (6)) are obtained in a data-driven manner based on training performance.
\begin{equation}
loss_{total} = \left(e^{-v_1} \cdot loss_1 + v_1\right) +\left(e^{-v_2} \cdot loss_2 + v_2\right)
\end{equation} 
Equation (6) is shown above, where the $loss_1$ and $loss_2$ represent the self-supervised contrastive loss and supervised contrastive loss, respectively. $v_1$ and $v_2$ denote the two weighting parameters that are automatically computed based on the training performance. Both two parameters are randomly initialized and included in the loss calculation per batch during the total loss optimization.} 

\begin{figure*}[t]
\begin{center}
\includegraphics[width=1\textwidth]{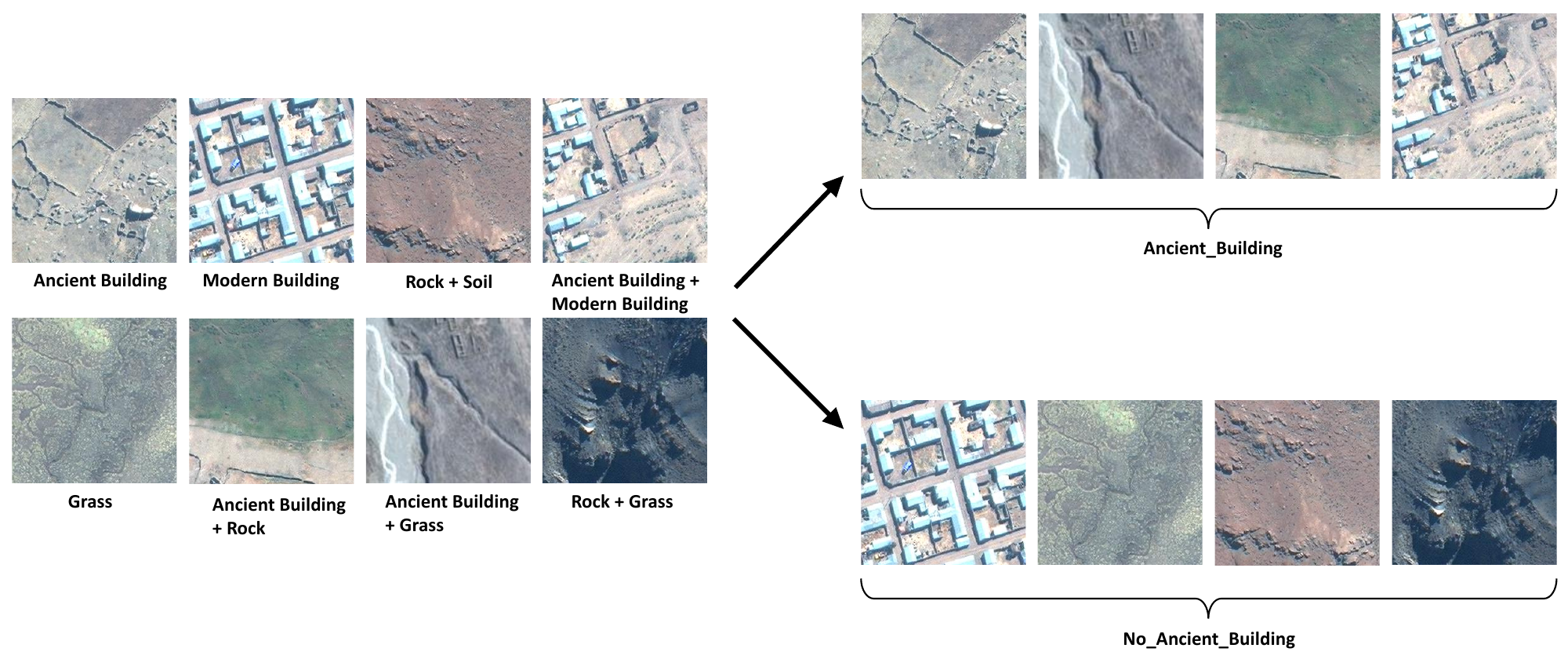}
\end{center}
\caption{\textbf{Example of annotated classes.} This figure demonstrates example classes of the annotated data. The left panel shows the various types of unannotated images with a mixture of contents, including ancient/modern buildings, rock, soil, and grass. Due to the variety of potential objects, it's unrealistic to create a separate class for each unique combination of objects. Therefore, two classes were created based on the presence of ancient buildings, which is shown in the right panel.}
\label{fig:classdesign}
\end{figure*}

\section{Data and Experiments}
\textcolor{black}{The data used in the experiments are collected from WorldView 2 and WorldView 3 satellite constellations. The Data subsection below provides the descriptions of data acquisition and the preprocessing pipeline. The Experiment Design subsection describes the experimental design, including data setup, hyperparameters, and validation metrics.} 

\subsection{Data}
The satellite images used in this analysis were collected by the WorldView 2 and WorldView 3 satellite constellations and were provided by the Digital Globe Foundation following color correction and orthographic correction using a coarse digital elevation model (DEM). The data was then pan-sharpened using the Bayesian fusion algorithm from the Orfeo-Toolbox~\citep{grizonnet2017orfeo} so as to increase the spatial resolution of the multi-spectral imagery to 0.5 m for the WorldView 2 imagery, and 0.3 m for the WorldView 3 imagery. In this study, all spatial bands were dropped except for the Red, Green, and Blue spectral bands, and the imagery was re-sampled from 32 bits to 8 bits in order to reduce the storage size and computational requirements. In total, the images covered approximately 12,000 km$^2$. Finally, the study region was divided into approximately 1.6 million image tiles of size 76.8$\times$76.8 meters (256$\times$256 pixels at 0.3 m resolution).

Due to the semi-arid environment and limited vegetation coverage of the south-central Andes, satellites are able to capture clear and unobscured images of the ground and of the archaeological features of interest. Furthermore, ancient structures in this region were primarily constructed from stone, leading to relatively good preservation and consequently, a high visibility in satellite imagery. Of the 1.6 million image tiles produced, 5,000 of were randomly selected and manually coded for the presence/absence of archaeological buildings. To better balance sample size for the sparsely distributed modern and ancient settlements on the landscape, an additional set of 830 images known to contain examples of archaeological or modern structures was added to provide additional representation for the aforementioned categories. 

Since ancient buildings were the objects of interest, all images were labeled into two classes: "ancient\_building" (that is, "the presence of an archaeological structure," defined as a human-made structures less than 30 m in its largest dimension without evidence of modern roofs or maintenance) and "no\_ancient\_building" (that is, "no presence of an archaeological structure"). From the remaining unlabeled images, around 100,000 images were randomly selected to train the self-supervised deep learning models. Finally, those images were visually examined, and the defective ones (missing data) were abandoned, resulting in an unlabeled dataset of 95,358 images. Sample images are presented in Figure~\ref{fig:classdesign}.

\subsection{Experimental Design}
\textbf{(1)} For the Semi-supervised contrastive pre-training: the training dataset consists of 95,358 unlabeled images and 258 labeled foreground images. \textbf{(2)} For the supervised downstream classification fine-tuning task: The 5,830 labeled images were divided into training, validation, and testing splits, ensuring that images from the nearby physical space were placed into the same split in order to avoid the issue of data contamination. Additionally, in order to alleviate the unbalanced nature of our data source, the under-representative positive class (with ancient builds) in our training dataset was up-sampled to have the roughly an equal distribution to the negative class. The details of the data split are shown in Table \ref{table:setup}. As a last step, all of the labeled and unlabeled images were resized to 128$\times$128 so as to expedite the training process.
\begin{table}[ht]
\centering
\tbl{Datasets Setup}
{\begin{tabular}{c|c|c}
\toprule
Dataset & \# of Ancient\_Building & \# of No\_Ancient\_Building \\
\midrule
Training & \begin{tabular}{@{}c@{}}193 (Original) \\ 3,088 (Upsampled)\end{tabular} & 4,272 \\
\midrule
Validation & 65 & 610  \\
\midrule
Testing & 71 & 619 \\
\midrule
Unannotated & \multicolumn{2}{c}{95,358} \\
\bottomrule
\end{tabular}}
\label{table:setup}
\end{table}
\subsubsection{Semi-Supervised Contrastive Training}
The proposed semi-supervised contrastive learning model was adopted from the SimSiam network with major modifications on the loss function. The optimizer used was the SGD optimizer which was initialized with a learning rate = 0.1, weight decay = 0.0001, and momentum = 0.9. The training dataset of the unlabeled images had a batch size = 512, while the labeled dataset had a batch size = 16. The Mixed-precision training features were integrated into our network in order to boost the training process. 

\textcolor{black}{The model was trained for 200 epochs with approximately 100 training hours. The 100 training hours were computed by a workstation with Intel Xeon Gold 5118 2.30 GHz CPU, 383 GB memory, and two NVIDIA GeForce RTX 2080 Ti GPU (11.0 GB dedicated GPU memory).} 

\subsubsection{Supervised Fine-Tuning and Testing}
After pre-training the model using the unannotated data, an additional single linear layer was fined tuned with the labeled data. \textcolor{black}{The F1 score and the balanced accuracy~\citep{wegier2020application, feng2021imbalanced} on the validation set were the metrics used to select the best performance epoch as well as the optimal hyper-parameters.} 

\subsubsection{Evaluation metrics}
According to~\citep{wegier2020application}, the F1 score aggregates the sensitivity and precision,  
\textcolor{black}{
\begin{equation}
F1\ score \ = 2\ \times \ \frac{Precision\ \times\ Sensitivity}{Precision\ + \ Sensitivity}
\end{equation}}
where the sensitivity (or recall) determines the accuracy of the minority class classification and precision indicates the probability of correct detection. 

Balanced accuracy is the arithmetic mean of the sensitivity and specificity,
\textcolor{black}{
\begin{equation}
    Balanced\ accuracy \ = \frac{Specificity\ + \ Sensitivity}{2}
\end{equation}}
The specificity, in a binary case, indicates the accuracy of recognizing the negative (majority) class. 

\section{Results}
\textcolor{black}{Extensive experiments are then designed to verify the effectiveness of our proposed model. Thus, a performance comparison with other state-of-art methods is presented for an ablation study.} 

\subsection{Ablation Study}
\textcolor{black}{This ablation study consisted of two alternative versions of our proposed pre-trained model. The first version conducted self-supervised contrastive learning using only loss\_cosine as discussed in the Methods section, inspired by the SimSiam network~\citep{chen2020simple}. By contrast, the second version conducted supervised contrastive learning using only loss\_super as discussed in the Methods section. The corresponding testing results on the downstream labeled data have been presented in Table~\ref{table:results} as SSL.}

\begin{table}[hb]
\tbl{Quantitaive results of different learning methods. SL and SSL correspond to Supervised Learning and Self-Supervised Learning, respectively. CE is short for Cross Entropy.}
{\begin{tabular}{c|c|c|c|c|}
\toprule
& Model & Loss Function & Balanced Accuracy & F1 Score\\
\midrule
 \multirow{2}{*}{SL}
 & ResNet 50\textsuperscript{a}
& Supervised CE Loss~\citep{zhang2018generalized}& 0.701 & 0.612 \\
& ResNet 50\textsuperscript{b}
& Supervised CE Loss~\citep{zhang2018generalized} & 0.790 & 0.718 \\

\midrule
\multirow{7}{*}{SSL}
& \multirow{2}{*}{SimCLR}
& Self-Supervised Loss~\citep{chen2020simple} & 0.766 & 0.554 \\
& & \textbf{Semi-Supervised Loss (Ours)} & \textbf{0.769} & \textbf{0.613}
\\ \cmidrule{2-5}

& \multirow{2}{*}{BYOL}
& Self-Supervised Loss~\citep{grill2020bootstrap} & 0.731 & 0.681 \\
& & \textbf{Semi-Supervised Loss (Ours)} & \textbf{0.757} & \textbf{0.696} \\ \cmidrule{2-5}
& \multirow{3}{*}{SimSiam}
& Self-Supervised Loss Only~\citep{chen2021exploring} & 0.752 & 0.744 \\
& & Supervised Loss Only~\citep{khosla2020supervised} & 0.500 & 0.477 \\
& & \textbf{Semi-Supervised Loss (Ours)} & \textbf{0.790} & \textbf{0.762} \\
\bottomrule
\end{tabular}}
\tabnote{\textsuperscript{a} This corresponds to model: ResNet 50 (from scratch)\\
 \textsuperscript{b} This corresponds to model: ResNet 50 (ImageNet pretrained)}
\label{table:results}
\end{table}

\subsection{Comparison with Fully-Supervised Learning Benchmark} 
\textcolor{black}{In addition to the contrastive learning framework discussed above, a fully supervised version of the experiment that was trained from scratch and used only the labeled images was also established, so as to further demonstrate the performance boost that is provided by the unlabeled data. Since our self-supervised framework employed ResNet50 as the backbone model, it's also used here for the canonical fully-supervised learning benchmarks. An SGD optimizer and cross entropy loss were employed to create a standard training environment. The model was trained for 16 epochs. The trained model with the best validation performance was selected to run the testing dataset. The results of investigating canonical fully-supervised benchmarks are demonstrated in Table~\ref{table:results} as SL.} 

\subsection{Experiments on Additional Contrastive Learning Frameworks}
\textcolor{black}{In order to further illustrate the effectiveness of our network design, two additional contrastive learning frameworks - BYOL and SimCLR - were utilized and modified to also incorporate the semi-supervised loss. The corresponding F1 score and balanced accuracy are indicated in Table~\ref{table:results}.} 
We then evaluated the accuracy of positive and negative images in this highly unbalanced scenario. Our semi-supervised loss mechanism yields an accuracy of 0.803 for positive images and 0.734 for negative images when using the SimCLR backbone. It suggested that our semi-supervised loss mechanism had a balanced performance for both positive and negative cases.

\section{Discussion}
In this study, we developed a new, semi-supervised contrastive learning pseudo-label generation method based on the similarity matrix; in doing so, our ultimate goal was to enhance self-supervised training performance over a highly unbalanced dataset. By integrating self-supervised and supervised loss functions, we designed a new learning framework that simultaneously learns from both unannotated and annotated data.

The conducted experiments have yielded promising results. In the ablation study, the models trained using only \textbf{loss\_super} were relatively ineffective in identifying ancient buildings, while the models trained using only \textbf{loss\_cosine} produced competitive F1 scores and balanced accuracies; this indicated that having a model pre-trained on unlabeled data using self-supervised contrastive learning was essential in distinguishing ancient buildings from other objects. Nonetheless, our proposed semi-supervised model that was trained with \textbf{loss\_cosine} + \textbf{loss\_super} outperformed its self-supervised and fully-supervised counterparts by a decent margin. This result showed that the integration of fully-supervised and self-supervised networks was a complimentary aggregation. Furthermore, when comparing fully-supervised learning benchmarks (such as ResNet50) with our model, our solution exhibited superior performance on the downstream labeled dataset.

\begin{figure}[t]
\begin{center}
\includegraphics[width=1\textwidth]{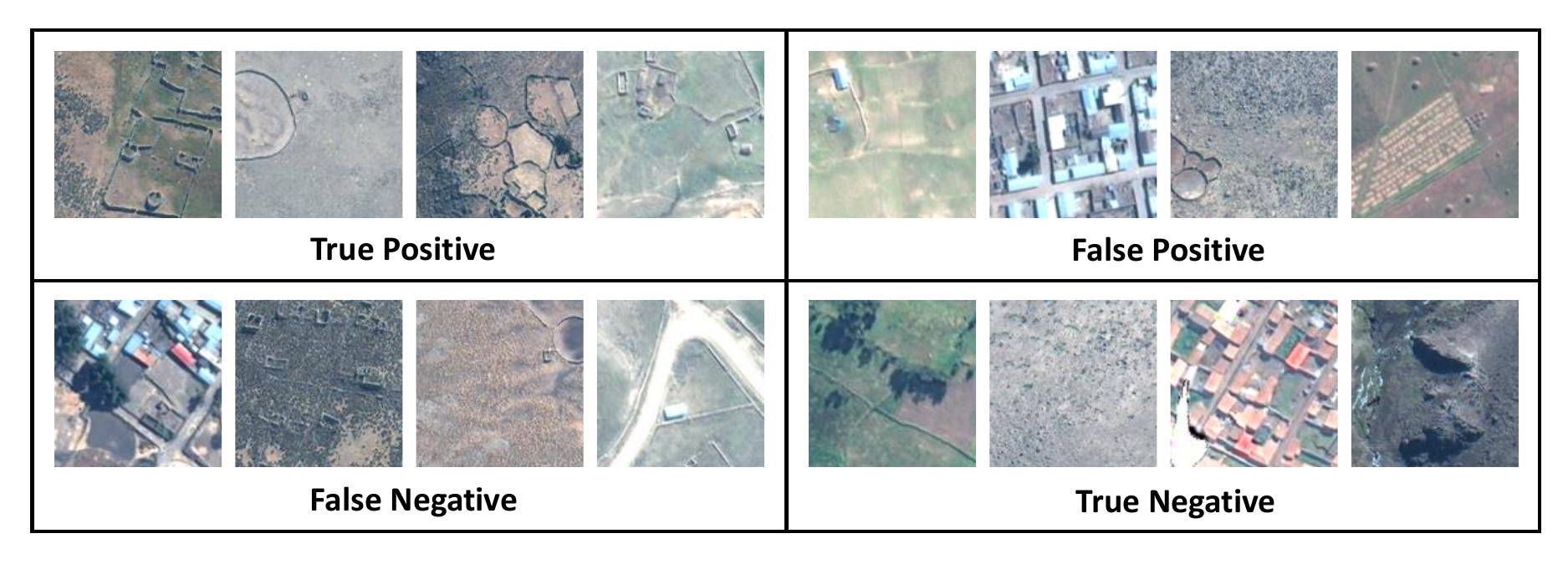}
\end{center}
\caption{\textbf{Testing Sample Results} This figure presents representative samples from the testing results. The left panel indicates the true positive examples while the right panel indicates the true negative cases. Likewise, the upper row indicates the predicted positive ones while the lower row indicates the predicted negative ones.}
\label{fig:confusionmatrix}
\end{figure}

Figure~\ref{fig:confusionmatrix} offered additional insights on the performance of our framework. The upper left and lower right corners indicate the correct classes prediction, while the upper right and lower left are the examples of false positives and false negatives. From the ablation study, the superior performance of our model can be attributed to the dynamic combination of fully-supervised and self-supervised information. The loss generated from the positive and pseudo-negative images serves as a complement to the other loss functions that were generated solely on unlabeled instances.

We believe there are several potential improvements for our semi-supervised contrastive learning framework. First, our proposed pseudo-labeling strategy would only work for highly unbalanced data sets. Moreover, the current model has not been extended to multi-label classification scenarios. Meanwhile, the size of annotated training images in our study was still relatively small. To further facilitate the performance, we might need more training data especially with more positive images (ancient building) from our predicted positive class.

\section{Conclusion}
In this project, we proposed a new semi-supervised contrastive learning method for identifying relict architectural features in the south-central Andes from satellite imagery. As opposed to the existing solutions, we utilized the unbalanced nature of the large-scale unlabeled data to form pseudo-negative pairs. Using such negative pairs, we leveraged the contrastive learning method by introducing a holistic learning scheme with both cosine similarity loss and pseudo-supervision loss functions. According to the experimental results, our proposed framework yielded both superior accuracy and a higher F1 score as compared with its self-supervised and fully-supervised counterparts. The integrated model eventually outperformed traditional supervised networks (e.g., ResNet50) by 15\% in F1 score and 8.9\% in balanced accuracy.

These improved feature detection results show great promise for developing a machine-human teaming approach, in which human surveyors would not need to visually scan vast featureless areas, and instead could focus their efforts on categorizing, annotating, and enriching the attribute data on autonomously-identified features. This approach would also eliminate inter-observer differences in feature detection sensitivity and specificity, while also enabling greater transparency and reproducibility through the reporting of model parameters.

\section*{Acknowledgements}
This work is supported by Scaling Success Grant from Vanderbilt University. The imagery analyzed in this paper was provided by the Digital Globe Foundation through a generous satellite imagery grant (S. Wernke, P.I.). Computational resources were provided by the Vanderbilt University Spatial Analysis Research Laboratory (https://wernkelab.org/). This work has not been submitted for publication or presentation elsewhere.

\section*{Data Availability Statement}
The data that support the findings of this study are available from the author, Steven A. Wernke, upon reasonable request.

\bibliographystyle{tfcad}
\bibliography{References}

\end{document}